\documentclass[conference]{IEEEtran}
\IEEEoverridecommandlockouts

\usepackage{cite}
\usepackage{graphicx}
\usepackage{amsmath,amssymb}
\usepackage{siunitx}
\usepackage{booktabs}
\usepackage{subfig}
\usepackage{xcolor}
\usepackage{url}
\usepackage{fancyhdr}
\usepackage{balance}
\usepackage[hidelinks]{hyperref}
\usepackage{caption}
\usepackage{subcaption}

\hypersetup{colorlinks=true,linkcolor=black,citecolor=black,urlcolor=black}

\begin{document}

\newcommand{\redHeart}{\includegraphics[height=1.2ex]{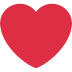}}
\newcommand{\fire}{\includegraphics[height=1.2ex]{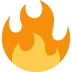}}
\newcommand{\faceWithTearsOfJoy}{\includegraphics[height=1.2ex]{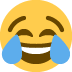}}
\newcommand{\jackoLantern}{\includegraphics[height=1.2ex]{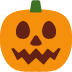}}
\newcommand{\foldedHands}{\includegraphics[height=1.2ex]{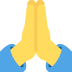}}
\newcommand{\rollingOnTheFloor}{\includegraphics[height=1.2ex]{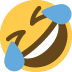}}
\newcommand{\sparkles}{\includegraphics[height=1.2ex]{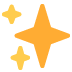}}
\newcommand{\maleSign}{\includegraphics[height=1.2ex]{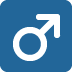}}
\newcommand{\femaleSign}{\includegraphics[height=1.2ex]{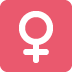}}
\newcommand{\loudlyCryingFace}{\includegraphics[height=1.2ex]{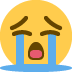}}
\newcommand{\coffee}{\includegraphics[height=1.2ex]{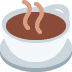}}
\newcommand{\lightSkinTone}{\includegraphics[height=1.2ex]{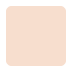}}
\newcommand{\thumbsUp}{\includegraphics[height=1.2ex]{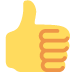}}
\newcommand{\ghost}{\includegraphics[height=1.2ex]{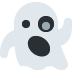}}
\newcommand{\faceBlowingAKiss}{\includegraphics[height=1.2ex]{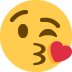}}
\newcommand{\smilingFaceWithHeartEyes}{\includegraphics[height=1.2ex]{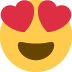}}
\newcommand{\rose}{\includegraphics[height=1.2ex]{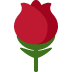}}
\newcommand{\smilingFaceWithHearts}{\includegraphics[height=1.2ex]{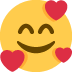}}
\newcommand{\clappingHands}{\includegraphics[height=1.2ex]{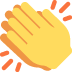}}
\newcommand{\partyPopper}{\includegraphics[height=1.2ex]{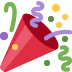}}

\newcommand{\northWestArrow}{\includegraphics[height=1.2ex]{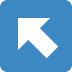}}
\newcommand{\upwardsBlackArrow}{\includegraphics[height=1.2ex]{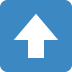}}
\newcommand{\southWestArrow}{\includegraphics[height=1.2ex]{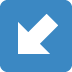}}
\newcommand{\leftwardsBlackArrow}{\includegraphics[height=1.2ex]{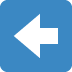}}
\newcommand{\rightwardsBlackArrow}{\includegraphics[height=1.2ex]{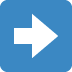}}
\newcommand{\downwardsBlackArrow}{\includegraphics[height=1.2ex]{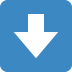}}
\newcommand{\speakerMediumVolume}{\includegraphics[height=1.2ex]{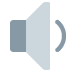}}
\newcommand{\speakerHighVolume}{\includegraphics[height=1.2ex]{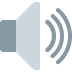}}
\newcommand{\northEastArrow}{\includegraphics[height=1.2ex]{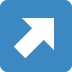}}
\newcommand{\southEastArrow}{\includegraphics[height=1.2ex]{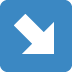}}
\newcommand{\fog}{\includegraphics[height=1.2ex]{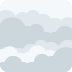}}
\newcommand{\neutralFace}{\includegraphics[height=1.2ex]{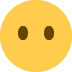}}
\newcommand{\japaneseCongratulations}{\includegraphics[height=1.2ex]{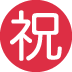}}
\newcommand{\onArrow}{\includegraphics[height=1.2ex]{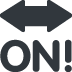}}
\newcommand{\topArrow}{\includegraphics[height=1.2ex]{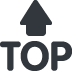}}

\newcommand{\blackLargeSquare}{\includegraphics[height=1.2ex]{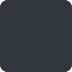}}
\newcommand{\cat}{\includegraphics[height=1.2ex]{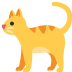}}
\newcommand{\spadeSuit}{\includegraphics[height=1.2ex]{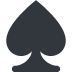}}
\newcommand{\diamondSuit}{\includegraphics[height=1.2ex]{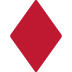}}
\newcommand{\backhandIndexLeft}{\includegraphics[height=1.2ex]{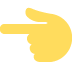}}
\newcommand{\backhandIndexRight}{\includegraphics[height=1.2ex]{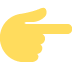}}
\newcommand{\pie}{\includegraphics[height=1.2ex]{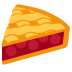}}
\newcommand{\cupcake}{\includegraphics[height=1.2ex]{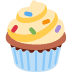}}
\newcommand{\clubSuit}{\includegraphics[height=1.2ex]{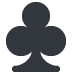}}
\newcommand{\heartSuit}{\includegraphics[height=1.2ex]{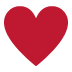}}
\newcommand{\candy}{\includegraphics[height=1.2ex]{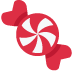}}
\newcommand{\lollipop}{\includegraphics[height=1.2ex]{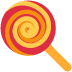}}
\newcommand{\chocolateBar}{\includegraphics[height=1.2ex]{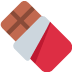}}
\newcommand{\softIceCream}{\includegraphics[height=1.2ex]{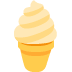}}
\newcommand{\iceCream}{\includegraphics[height=1.2ex]{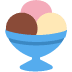}}
\newcommand{\whiteCircle}{\includegraphics[height=1.2ex]{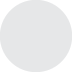}}
\newcommand{\redCircle}{\includegraphics[height=1.2ex]{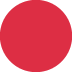}}
\newcommand{\balloon}{\includegraphics[height=1.2ex]{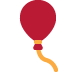}}
\newcommand{\mapleLeaf}{\includegraphics[height=1.2ex]{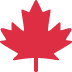}}
\newcommand{\fallenLeaf}{\includegraphics[height=1.2ex]{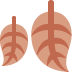}}
\newcommand{\sunrise}{\includegraphics[height=1.2ex]{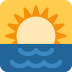}}
\newcommand{\openHands}{\includegraphics[height=1.2ex]{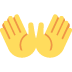}}
\newcommand{\musicalNote}{\includegraphics[height=1.2ex]{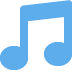}}
\newcommand{\musicalNotes}{\includegraphics[height=1.2ex]{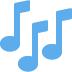}}
\newcommand{\wrappedGift}{\includegraphics[height=1.2ex]{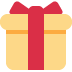}}
\newcommand{\birthdayCake}{\includegraphics[height=1.2ex]{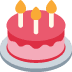}}

\newcommand{\rollingOnFloorLaughing}{\includegraphics[height=1.2ex]{emojis/1f923.png}}
\newcommand{\grinningSquintingFace}{\includegraphics[height=1.2ex]{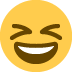}}
\newcommand{\poutingFace}{\includegraphics[height=1.2ex]{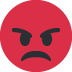}}
\newcommand{\flushedFace}{\includegraphics[height=1.2ex]{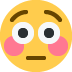}}
\newcommand{\wearyFace}{\includegraphics[height=1.2ex]{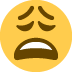}}
\newcommand{\faceWithSteamFromNose}{\includegraphics[height=1.2ex]{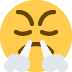}}
\newcommand{\pleadingFace}{\includegraphics[height=1.2ex]{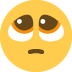}}
\newcommand{\faceWithOpenMouth}{\includegraphics[height=1.2ex]{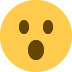}}
\newcommand{\angryFace}{\includegraphics[height=1.2ex]{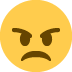}}
\newcommand{\perseveringFace}{\includegraphics[height=1.2ex]{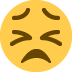}}
\newcommand{\faceWithSymbolsOnMouth}{\includegraphics[height=1.2ex]{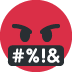}}
\newcommand{\explodingHead}{\includegraphics[height=1.2ex]{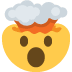}}
\newcommand{\blueHeart}{\includegraphics[height=1.2ex]{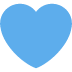}}
\newcommand{\confettiBall}{\includegraphics[height=1.2ex]{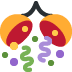}}
\newcommand{\doughnut}{\includegraphics[height=1.2ex]{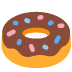}}
\newcommand{\hotBeverage}{\includegraphics[height=1.2ex]{emojis/2615.png}}
\newcommand{\cookie}{\includegraphics[height=1.2ex]{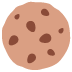}}
\newcommand{\usaFlag}{\includegraphics[height=1.2ex]{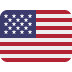}}

\title{Emoji–Emotion Ranking System Using Twitter Data}


\author{
\IEEEauthorblockN{Danila Khlebokazov}
\IEEEauthorblockA{School of Information Technology \\ and Engineering, KBTU\\
Almaty, Kazakhstan\\
khlebokazov\_danila@mail.ru}
\and
\IEEEauthorblockN{Nurkhan Tashimov}
\IEEEauthorblockA{School of Information Technology\\ and Engineering, KBTU\\
Almaty, Kazakhstan\\
tashimov\_nurh@mail.ru}
\and
\IEEEauthorblockN{Pakizar Shamoi}
\IEEEauthorblockA{School of Information Technology\\ and Engineering, KBTU\\
Almaty, Kazakhstan\\
p.shamoi@kbtu.kz}
}

\fancypagestyle{firstpagestyle}{
    \fancyhf{} 
    \fancyhead[L]{2026 IEEE 5th International Conference on Smart Information Systems and Technologies (SIST) \\ 13-15 May, 2026, Astana, Kazakhstan}
    \renewcommand{\headrulewidth}{0pt} 
    \renewcommand{\footrulewidth}{0pt} 
}

\maketitle
\thispagestyle{firstpagestyle}

\begin{abstract}

Nowadays, emojis are often replacing words. Yet computational systems still oversimplify them.
Most existing approaches treat emojis as static sentiment indicators and overlook their emotional distributions. In this study, we propose an emoji-aware emotion analysis framework based on a Twitter (X) dataset of 100,000 emoji-containing replies collected between 2020 and 2025. After preprocessing and text cleaning, we applied text-to-emotion classification to detect five primary emotions (Happy, Angry, Sad, Fear, and Surprise) for each message. By aggregating emotion scores across contexts in which each emoji appears, we estimate emoji–emotion association distributions and construct an emoji–emotion ranking system reflecting relative emotional dominance. Furthermore, we project emojis into the Russell valence–arousal space to enable continuous affective interpretation. Our results demonstrate that emojis exhibit probabilistic, context-sensitive emotional profiles rather than fixed sentiment polarities.

\end{abstract}


\section{Introduction}


Emojis are now central to how people express affect online, often shifting or even inverting the sentiment implied by text alone \cite{Caspi2024}. 

Emojis have transformed digital communication, becoming among the most widely used affective signals in online interactions \cite{Huh2025}. Across messaging platforms and social media, they function as compact carriers of emotion, tone, and interpersonal nuance. For younger generations in particular, emojis are deeply embedded in everyday life and communication, often replacing or modifying text \cite{2025The, Zaheer2025Emojis,  Ali-Chand2024Exploring}. They can convey irony, affection, sarcasm, excitement, or frustration. In many cases, the presence or absence of a single emoji can substantially alter the perceived emotional meaning of a message \cite{Holtgraves2023, Caspi2024}.

The growing reliance on emojis presents a significant challenge for computational systems tasked with sentiment and emotion analysis. Despite their prevalence, most existing approaches treat emojis as static sentiment markers, assigning fixed polarity labels or assuming stable emotional meaning \cite{Al-Azani2021Early, Gavilanes2018, Mabokela2023Investigating, Kralj2015emojis}. Such simplifications overlook the contextual variability of emoji use and risk misinterpreting affective intent in informal, youth-dominated digital discourse. As emojis increasingly shape online expression, there is a need for modeling approaches that capture their multidimensional, context-dependent emotional characteristics.

This paper presents a framework for constructing emoji–emotion distributions and ranking emojis according to dominant emotional tendencies, based on real-world Twitter usage. 
In addition, we embed emojis in a continuous affective space (Russell) to enable analysis and visualization.

The main contributions of this work are summarized as follows:

\begin{itemize}
    \item Collection of a dataset of 100{,}000 emoji-containing Twitter (X) replies collected between 2020 and 2025

    \item Development of a context-aggregated emoji--emotion modeling approach in which each tweet is labeled with weighted emotion scores (Happy, Angry, Surprise, Sad, Fear)

    \item Design of an emoji--emotion ranking mechanism that identifies dominant emotional tendencies for each emoji and supports comparative analysis across emojis.

\end{itemize}

Section 1 is this Introduction. Section II reviews related work. Section III describes methods, including the construction of the dataset. Section IV reports experiments and results, including emoji Russell mapping.  Finally, Section V provides concluding remarks.

\section{Related Work}
This section provides an overview of emoji-based emotion detection approaches. 

The integration of emojis into natural language processing (NLP) has transformed sentiment and emotion detection, especially in social media and informal digital communication \cite{Grover2026}. Emojis serve as visual cues that can reinforce, invert, or clarify the emotional tone of text \cite{Neel2023, Cabajar2026}. Recent NLP research treats emojis as labels, features, or an additional modality to boost sentiment and emotion detection across languages, domains, and platforms.

Emoji processing in affective computing has been approached via distant supervision, in which emojis are treated as weak labels for large-scale learning \cite{Liu2021, Felbo2017, Chen2021, Singh2024}. By predicting emojis from large Twitter corpora, neural models learn transferable affect representations that improve the detection of sentiment, emotion, and sarcasm \cite{Tomihira2020}. However, this strategy often assumes relatively stable emoji–emotion associations and does not explicitly model multidimensional emotional distributions.

Another direction is to construct emoji sentiment lexicons using unsupervised methods based on co-occurrence statistics or external resources such as Emojipedia. These lexica assign polarity scores to emojis and enhance traditional sentiment analysis pipelines \cite{Gavilanes2018, Alfreihat2024}. 


More recent studies combine text and emoji as multimodal inputs by fusing transformer-based text embeddings with emoji embeddings through attention mechanisms.  Hybrid and multi‑view models fuse text embeddings (often BERT) with explicit emoji features or emoji embeddings \cite{Kuruva2025, Liu2021Improving}, using attention or cross‑attention to learn complementary signals \cite{Khan2025, Tang2024, Gupta2023, Kusal2025, Kuruva2025, Xu2024}.

Emoji-aware models have been successfully applied to English-Hindi code-mixed texts \cite{Singh2024}, Chinese Weibo posts \cite{Liu2021Improving}, Arabic tweets \cite{Alfreihat2024}, Japanese/English multilingual corpora \cite{Tomihira2020}, customer reviews (aviation sector) \cite{Khan2025}, marketing analytics \cite{Hotz-Behofsits2025EXPRESS:}, mental health monitoring \cite{Joseph2024Emotion}, software engineering communication \cite{Chen2021}, working chats \cite{igali2024trackingemotionaldynamicschat} and more.

Despite these advances, challenges remain, such as handling polysemous/multifunctional emojis, addressing domain adaptation, and improving the explainability of deep models’ decisions in emoji-text interactions.














\section{Methods}

\subsection{Data}

We constructed the dataset from 100,000 publicly available Twitter (X) replies collected between January 1, 2020, and November 1, 2025. The collection process was implemented as an automated search-based pipeline. For each target emoji, a dedicated Twitter (X) search query was issued over the selected time interval, with filters restricting results to English-language replies and excluding posts containing links. This design was intended to focus the dataset on short, conversational, emoji-containing responses that are more suitable for contextual emotion analysis.

To initialize the search space, we used a predefined list of target emojis derived from the Emoji Sentiment Ranking v1.0 dataset \cite{Kralj2015emojis}, which contains 614 annotated emojis. Not all emojis from this resource were observed in the collected Twitter (X) data. After applying the filtering and consolidation steps, the resulting corpus contained texts with 531 of the 614 emojis from Emoji Sentiment Ranking v1.0. In addition, the final corpus included emojis not present in the original ranking resource. Overall, the final dataset consists of 100,000 replies containing 1,393 distinct emojis.

The collection pipeline consists of the following stages:
\begin{itemize}
  \item \textbf{Query generation.} A predefined list of target emojis was used to construct search queries over Twitter (X). Each query combined a target emoji with temporal and linguistic constraints covering the period from January 1, 2020, to November 1, 2025.
  \item \textbf{Reply-oriented filtering.} The search configuration retained only replies in English and excluded posts containing links, thereby favoring short conversational texts with direct emoji usage.
  \item \textbf{Record extraction.} For each matched post, the collected record included the query emoji, the post URL, and the visible textual content.
  \item \textbf{Per-query storage.} Retrieved records were first stored in partitioned per-emoji files in order to support scalable collection and later merging.
  \item \textbf{Deduplication and merging.} Duplicate instances were filtered both during collection and during the final consolidation step, after which all partitioned files were merged into a single combined dataset.
\end{itemize}

During collection, duplicate posts were identified using post URLs, and repeated rows were removed again when all partitioned files were merged into the final dataset. The resulting combined dataset stored three primary fields for each instance: the target emoji, the post URL, and the corresponding text. This two-stage deduplication procedure reduced redundancy and helped ensure that the final corpus better reflected distinct emoji-use contexts.

\begin{figure}[!t]
  \centering
  \includegraphics[width=0.95\linewidth]{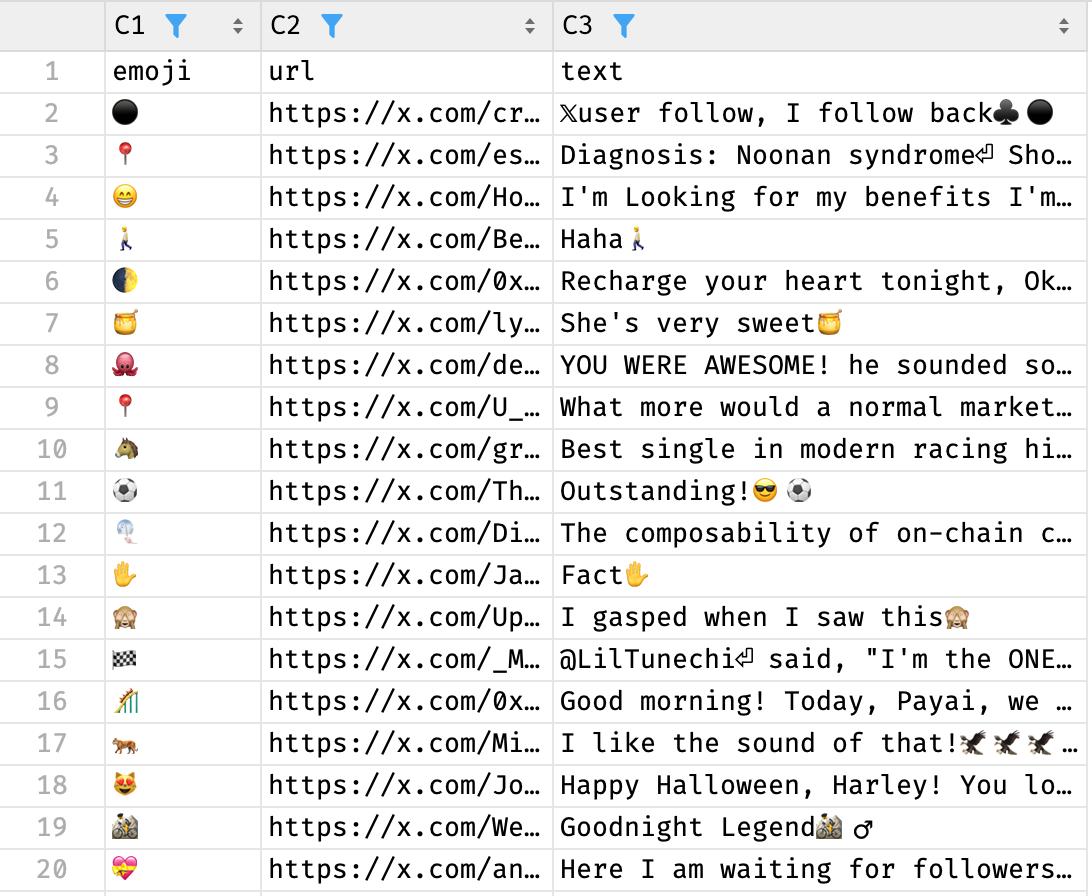}
  \caption{Samples from a collected Twitter (X) dataset.}
  \label{fig:dataset_example}
\end{figure}

Fig. \ref{fig:dataset_example} shows a sample of Twitter (X) replies containing emojis, illustrating the type of texts that were collected for our dataset. 


\begin{table}[!t]
  \centering
  \caption{Descriptive statistics of the dataset}
  \label{tab:stats}
  \begin{tabular}{lrr}
    \toprule
    \textbf{Field} & \textbf{Count} & \textbf{Share (\%)} \\
    \midrule
    Total posts & 100{,}000 & 100.0 \\
    With $\geq$2 emojis & 14{,}646 & 14.6 \\
    Avg. tokens / emojis & 11.94 / 1.29 & – \\
    \bottomrule
  \end{tabular}
\end{table}

Table \ref{tab:stats} shows the main characteristics of the collected Twitter (X) dataset. Out of 100,000 posts, 14,646 posts (14.6 \%) contain two or more emojis. On average, each post contains 11.94 tokens and 1.29 emojis, indicating that while most posts include only one emoji, a substantial portion includes multiple emojis.

\subsection{Text Emotion Detection}

To estimate the emotional content of the collected texts, we employed the \texttt{text2emotion} Python library \cite{text2emotion_github}, which maps text into five primary emotion categories: \textit{Happy, Angry, Surprise, Sad,} and \textit{Fear}. The method was selected as a lightweight and interpretable lexicon-based approach suitable for large-scale corpus analysis. In contrast to transformer-based models, it does not require manually labeled training data or additional model fine-tuning, which makes it easier to reproduce and apply consistently across a large collection of social media posts.

The choice of this method was aligned with the purpose of the present study. Our goal was not to develop or benchmark a state-of-the-art tweet-level emotion classifier, but rather to obtain an approximate contextual emotion signal that could be aggregated across many posts for emoji-level profiling. In this setting, the output of \texttt{text2emotion} is treated as an estimate of the emotional context in which an emoji is used, rather than as a definitive label of the emoji’s inherent meaning.

For each text, the library outputs a normalized score for each emotion, indicating the relative intensity of that emotion in the text. These scores were subsequently associated with the emojis present in the corresponding post. Because the final analysis aggregates data from many posts that contain the same emoji, the method is appropriate for corpus-level emoji profiling, even if individual post-level predictions are imperfect.

Before emotion extraction, the textual data were cleaned by removing URLs, user mentions, and hashtags to reduce platform-specific noise and improve the quality of the lexical signal for emotion estimation.

At the same time, we acknowledge the limitations of this approach. Lexicon-based emotion detection may be less reliable for short, noisy social media texts, especially when sarcasm, irony, implicit affect, or non-standard language is present. In addition, the detected emotions reflect the contextual emotional tone of the text in which an emoji appears, not the intrinsic meaning of the emoji itself. Therefore, the resulting scores should be interpreted as approximate corpus-level signals that become more informative after aggregation, rather than as exact emotion annotations for individual posts.

\subsubsection{Evaluation Criteria}

Our evaluation is descriptive and consistency‑oriented, aligned with the goal of building corpus‑level emoji–emotion profiles.

\textbf{Tweet‑level sanity.} We inspect random samples of replies after preprocessing to verify that detected emotions are plausible for the surrounding text and that obvious artifacts (URLs, mentions, hashtags) are removed.

\textbf{Emoji‑level stability.} We recompute emoji profiles on random halves of the corpus and report cosine similarity between per‑emoji emotion vectors, as well as agreement of the dominant emotion per emoji.

\textbf{Valence–Arousal coherence.} We check that emojis associated with positive/high‑energy texts tend to map to higher valence/arousal regions in the Russell space and that clearly negative emojis fall into low‑valence areas.

\subsubsection{Limitations}

\textbf{No human labels.} We do not use external human annotations; results are exploratory and rely on aggregation over many posts.

\textbf{Lexicon sensitivity.} Short, noisy, sarcastic, or code‑mixed texts can mislead a lexicon detector; aggregation reduces but does not remove this effect.

\textbf{Context, not essence.} Scores reflect the emotion of the surrounding text, not a fixed, universal meaning of the emoji; profiles may vary by domain and time.

\textbf{Scope.} English Twitter/X replies from 2020–2025 only; generalization to other languages/platforms is limited.

\textbf{Composition.} Multiple emojis in one post can blend/contrast affects; simple aggregation under‑models such interactions.

\textbf{Future work.} Adding a small human‑labeled subset and a contextual neural baseline would strengthen validation; we leave this for follow‑up work.

\subsection{Russell Mapping for Emojis}
\begin{figure}[!t]
  \centering
  \includegraphics[width=1\linewidth]{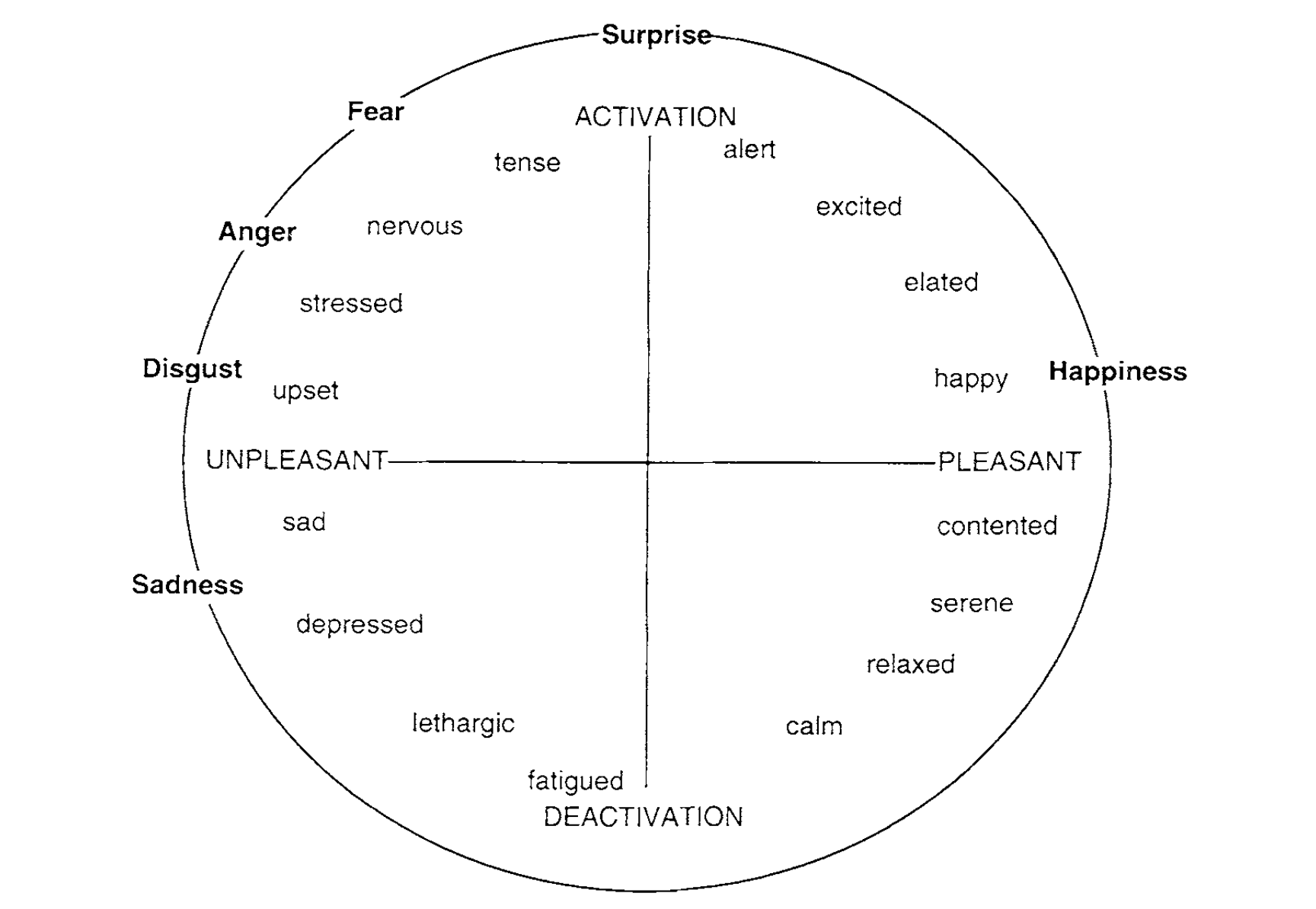}
  \caption{Russell Circumplex Model of Affect illustrating the two-dimensional structure of emotions defined by Valence (horizontal axis) and Arousal (vertical axis).}
  \label{fig:russell_and_feldman_emotions}
\end{figure}

To further analyze the emotional content of emojis, we mapped the detected emotions onto the Russell Circumplex Model of Affect \cite{Russell1999}. This model organizes emotions along two continuous dimensions: \textit{valence} (pleasant–unpleasant) and \textit{arousal} (activation–deactivation). By positioning emotions within this two-dimensional space, it becomes possible to visualize and systematically compare the emotional characteristics of different emojis.

Fig. \ref{fig:russell_and_feldman_emotions} illustrates the Russell Emotion Map, showing the spatial arrangement of the primary emotion categories used.

\subsection{Proposed Approach}
\begin{figure*}
    \centering
    \includegraphics[width=0.8\linewidth]{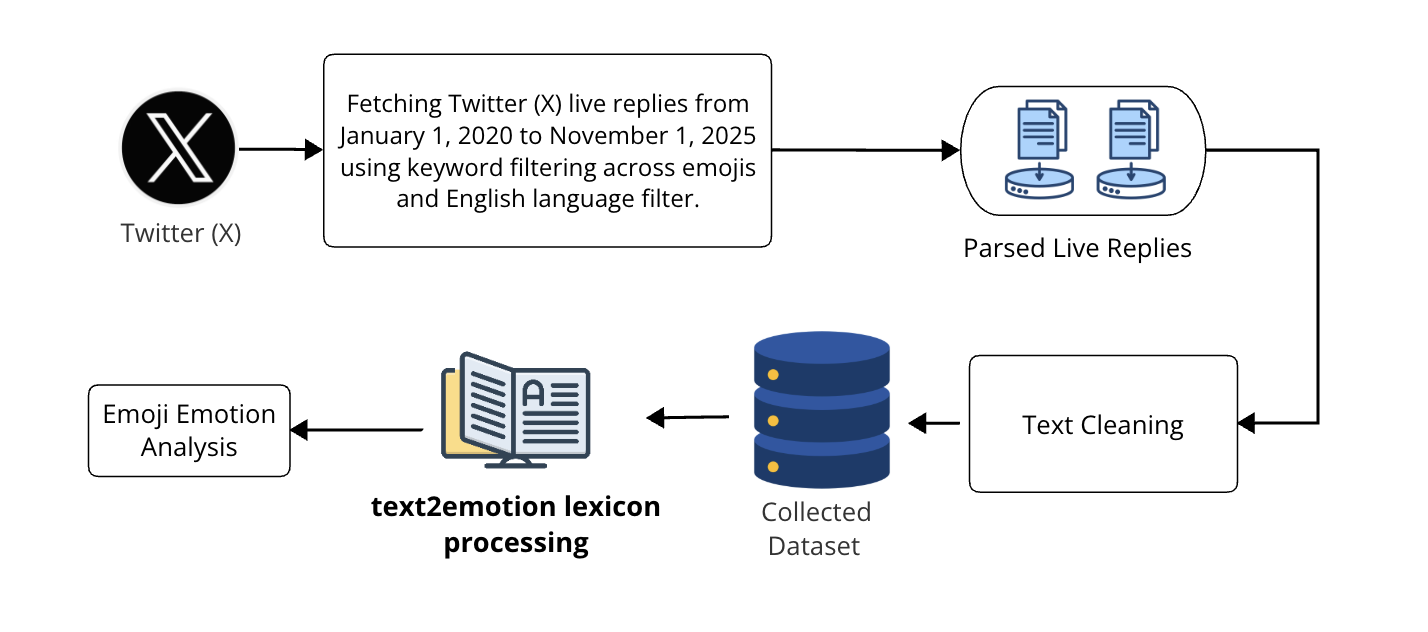}
    \caption{High-level view of the proposed approach}
    \label{fig:methodology}
\end{figure*}
The pipeline is presented in Fig. \ref{fig:methodology}. The process consists of sequential stages for collecting, preprocessing, and analyzing emoji-containing texts from Twitter (X).

\subsubsection{Data Collection} Publicly available Twitter (X) replies were retrieved from the search interface in live search mode for the period between January 1, 2020, and November 1, 2025. Data collection was performed using emoji-specific search queries over the Twitter (X) search interface, restricted to English-language replies and excluding posts containing links. This procedure resulted in a structured corpus of emoji-containing replies.

\subsubsection{Parsing and Preprocessing} The fetched live replies were parsed to extract relevant textual content and metadata. A text-cleaning procedure was then applied, including the removal of URLs, user mentions, hashtags, and non-textual artifacts. The cleaned texts were tokenized to separate word and emoji tokens. The processed texts were subsequently compiled into the final analytical dataset.

\subsubsection{Emotion Detection via Lexicon-Based Processing} The cleaned textual data were analyzed using the \texttt{text2emotion} Python library, which applies a lexicon-based approach to estimate emotional content. For each text, the model produces normalized scores across five primary emotions: \textit{Happy}, \textit{Angry}, \textit{Surprise}, \textit{Sad}, and \textit{Fear}. These scores represent the proportion of each emotion in the text.

\subsubsection{Emoji–Emotion Analysis} The emotion intensity obtained from the textual analysis was then associated with the emojis present in each corresponding post. By aggregating emotion intensity across posts containing the same emoji, an emoji-level emotional profile was constructed. 

\subsubsection{Data Reproducibility}
The study supports partial reproducibility through procedural transparency and detailed reconstruction of the collection pipeline. In particular, the manuscript documents the collection window, the use of emoji-specific search queries, the English-language and reply-oriented filtering criteria, the exclusion of posts containing links, the deduplication procedure, and the preprocessing steps applied prior to emotion analysis. In addition, the structure of the resulting dataset is described explicitly in terms of its main fields and final corpus size. While exact reconstruction may depend on platform dynamics and data availability over time, these methodological details are intended to make the dataset-building process transparent and reproducible at the procedural level.

\section{Results}

\begin{table}[ht]
\centering
\caption{Top 20 most frequent emojis}
\label{tab:top20_emojis}
\begin{tabular}{l S[table-format=6.4]}
\toprule
\textbf{Emoji} & \textbf{Count} \\
\midrule
\redHeart                    & 47059 \\
\fire                        & 40932 \\
\jackoLantern                & 34494 \\
\faceWithTearsOfJoy          & 31543 \\
\foldedHands                 & 29359 \\
\rollingOnTheFloor           & 25216 \\
\sparkles                    & 23671 \\
\maleSign                    & 23475 \\
\femaleSign                  & 20230 \\
\loudlyCryingFace            & 20225 \\
\coffee                      & 18541 \\
\lightSkinTone               & 18121 \\
\thumbsUp                    & 16870 \\
\ghost                       & 16618 \\
\faceBlowingAKiss            & 16406 \\
\smilingFaceWithHeartEyes    & 15505 \\
\rose                        & 14731 \\
\smilingFaceWithHearts       & 14319 \\
\clappingHands               & 14191 \\
\partyPopper                 & 14041 \\
\bottomrule
\end{tabular}
\end{table}

Table~\ref{tab:top20_emojis} confirms a heavy‑tailed usage pattern in which a small set of emojis accounts for a large share of posts. The leaders are \redHeart~(47{,}059) and \fire~(40{,}932), followed by \jackoLantern~(34{,}494), \faceWithTearsOfJoy~(31{,}543), and \foldedHands~(29{,}359). Such head concentration implies that emotion profiles for high‑ranked emojis are estimated on abundant context and are therefore more stable, whereas long‑tail emojis require caution due to data sparsity. The mix of affiliative/celebratory symbols (e.g., hearts, party popper) with seasonal icons (e.g., \jackoLantern) also hints at topic‑ and calendar‑driven effects that can shift distributions over time. In downstream use, we recommend prioritizing head emojis for reliable signals and treating tail emojis as exploratory.


\begin{table}[ht]
\centering
\caption{Top 20 most probable combination}
\label{tab:top20_most_probable_combination}
\begin{tabular}{lcccc}
\toprule
\textbf{Emoji 1} & \textbf{Emoji 2} & \textbf{Count} & \textbf{Probability} \\
\midrule
\fog                 & \neutralFace      & 129 & 0.8836 \\
\blackLargeSquare    & \cat              & 418 & 0.7902 \\
\spadeSuit           & \diamondSuit      & 157 & 0.7850 \\
\backhandIndexLeft   & \backhandIndexRight & 250 & 0.7485 \\
\pie                 & \cupcake          & 143 & 0.7333 \\
\spadeSuit           & \clubSuit         & 145 & 0.7250 \\
\spadeSuit           & \heartSuit        & 143 & 0.7150 \\
\candy               & \lollipop         & 723 & 0.7081 \\
\clubSuit            & \diamondSuit      & 157 & 0.7072 \\
\chocolateBar        & \lollipop         & 517 & 0.6949 \\
\chocolateBar        & \candy            & 499 & 0.6707 \\
\softIceCream        & \iceCream         & 158 & 0.6295 \\
\candy               & \jackoLantern     & 633 & 0.6200 \\
\whiteCircle         & \redCircle        & 252 & 0.6146 \\
\lollipop            & \jackoLantern     & 593 & 0.6095 \\
\balloon             & \partyPopper      & 420 & 0.6078 \\
\mapleLeaf           & \fallenLeaf       & 465 & 0.5762 \\
\sunrise             & \openHands        & 317 & 0.5611 \\
\musicalNote         & \musicalNotes     & 331 & 0.5535 \\
\wrappedGift         & \birthdayCake     & 390 & 0.5524 \\
\bottomrule
\end{tabular}
\end{table}

Table~\ref{tab:top20_most_probable_combination} highlights pairs with high conditional probability of the second emoji given the first, revealing tight thematic “bundles.” Sweets form a coherent cluster (\lollipop–\candy, $p{=}0.7081$; \chocolateBar–\lollipop, $p{=}0.6949$; \chocolateBar–\candy, $p{=}0.6707$), while card suits often co‑appear (\spadeSuit–\diamondSuit, $p{=}0.7850$; \clubSuit–\diamondSuit, $p{=}0.7072$). Seasonal motifs are strong (e.g., \candy–\jackoLantern, $p{=}0.6200$), suggesting that context is frequently set by events and holidays rather than pure sentiment. High conditionality indicates compositional meaning: the effect of a pair is not always a simple sum of its parts. Applications that interpret multi‑emoji posts should therefore reason over pairs (or short sequences), not just single‑emoji priors.

\begin{table}[ht]
\centering
\caption{Unexpected Emotions}
\label{tab:unexpected_emotions}
\begin{tabular}{lcccc}
\toprule
\textbf{Emoji} & \textbf{Expected} & \textbf{Actual} & \textbf{Exp. Intensity} & \textbf{Act. Intensity} \\
\midrule
\faceWithTearsOfJoy        & Happy & Fear & 0.2039 & 0.2978 \\
\rollingOnFloorLaughing    & Happy & Fear & 0.1978 & 0.2999 \\
\loudlyCryingFace          & Sad & Fear & 0.2442 & 0.2651 \\
\grinningSquintingFace     & Happy & Fear & 0.2425 & 0.2609 \\
\poutingFace               & Angry & Fear & 0.1172 & 0.2966 \\
\flushedFace               & Surprise & Fear & 0.2698 & 0.2719 \\
\wearyFace                 & Sad & Fear & 0.2722 & 0.2800 \\
\faceWithSteamFromNose     & Angry & Fear & 0.0966 & 0.3039 \\
\pleadingFace              & Sad & Happy & 0.2131 & 0.3582 \\
\faceWithOpenMouth         & Surprise & Fear & 0.2517 & 0.2667 \\
\angryFace                 & Angry & Fear & 0.1269 & 0.3068 \\
\perseveringFace           & Sad & Fear & 0.2747 & 0.2764 \\
\faceWithSymbolsOnMouth    & Angry & Fear & 0.1080 & 0.3033 \\
\explodingHead             & Surprise & Fear & 0.2833 & 0.3097 \\
\bottomrule
\end{tabular}
\end{table}

Next, Table~\ref{tab:unexpected_emotions} surfaces emojis whose corpus‑dominant emotion diverges from conventional expectations, underscoring context sensitivity. Laughter‑related emojis such as \faceWithTearsOfJoy and \rollingOnFloorLaughing, typically read as \textit{Happy} (expected 0.2039 and 0.1978), skew toward \textit{Fear} in our data (0.2978 and 0.2999), likely reflecting usage in anxious, ironic, or tension‑filled threads. Conversely, \pleadingFace  shows a shift toward \textit{Happy} (0.3582) despite an expected \textit{Sad} signal (0.2131), pointing to supportive or affectionate contexts. These reversals caution against fixed polarity labels and motivate the use of per‑corpus distributions and top‑emotion margins when ranking emojis for sentiment tasks. We provide these cases as diagnostic anchors for error analysis and system calibration.

\begin{table}[ht]
\centering
\caption{Top 20 emojis by co-occurrence}
\label{tab:top20_cooccurrence}
\begin{tabular}{lcccc}
\toprule
\textbf{Emoji 1} & \textbf{Emoji 2} & \textbf{Count} & \textbf{Probability} \\
\midrule
\jackoLantern        & \ghost              & 1364 & 0.4219 \\
\lollipop            & \candy              & 723 & 0.7081 \\
\candy               & \jackoLantern       & 633 & 0.6200 \\
\lollipop            & \jackoLantern       & 593 & 0.6095 \\
\redHeart            & \fire               & 521 & 0.1344 \\
\chocolateBar        & \lollipop           & 517 & 0.6949 \\
\chocolateBar        & \candy              & 499 & 0.6707 \\
\birthdayCake        & \partyPopper        & 493 & 0.4612 \\
\blueHeart           & \foldedHands        & 481 & 0.1241 \\
\mapleLeaf           & \fallenLeaf         & 465 & 0.5762 \\
\confettiBall        & \partyPopper        & 429 & 0.3349 \\
\balloon             & \partyPopper        & 420 & 0.6078 \\
\blackLargeSquare    & \cat                & 418 & 0.7902 \\
\doughnut            & \balloon            & 415 & 0.4183 \\
\hotBeverage         & \cookie             & 406 & 0.3798 \\
\chocolateBar        & \wrappedGift        & 406 & 0.2043 \\
\candy               & \ghost              & 397 & 0.5336 \\
\blueHeart           & \jackoLantern       & 396 & 0.3879 \\
\blueHeart           & \pleadingFace       & 395 & 0.1019 \\
\blueHeart           & \smilingFaceWithHearts & 393 & 0.1014 \\
\bottomrule
\end{tabular}
\end{table}

Table~\ref{tab:top20_cooccurrence} lists frequent co‑occurrences that shape message‑level affect beyond single‑emoji cues. The strongest seasonal pair is \jackoLantern–\ghost (1{,}364), while celebration and sweets pairs (e.g., \balloon–\partyPopper, $p{=}0.6078$; \birthdayCake–\partyPopper, 493/$p{=}0.4612$) cluster around positive, high‑arousal contexts. Mixed‑valence neighborhoods (e.g., heart colors with non‑affiliative symbols) indicate that users often combine emojis to nuance tone or signal irony. For practical use, we recommend (i) leveraging co‑occurrence neighborhoods to disambiguate ambiguous emojis, and (ii) down‑weighting single‑emoji priors when strong co‑occurring partners are present.



\begin{figure}
    \centering
    \includegraphics[width=1\linewidth]{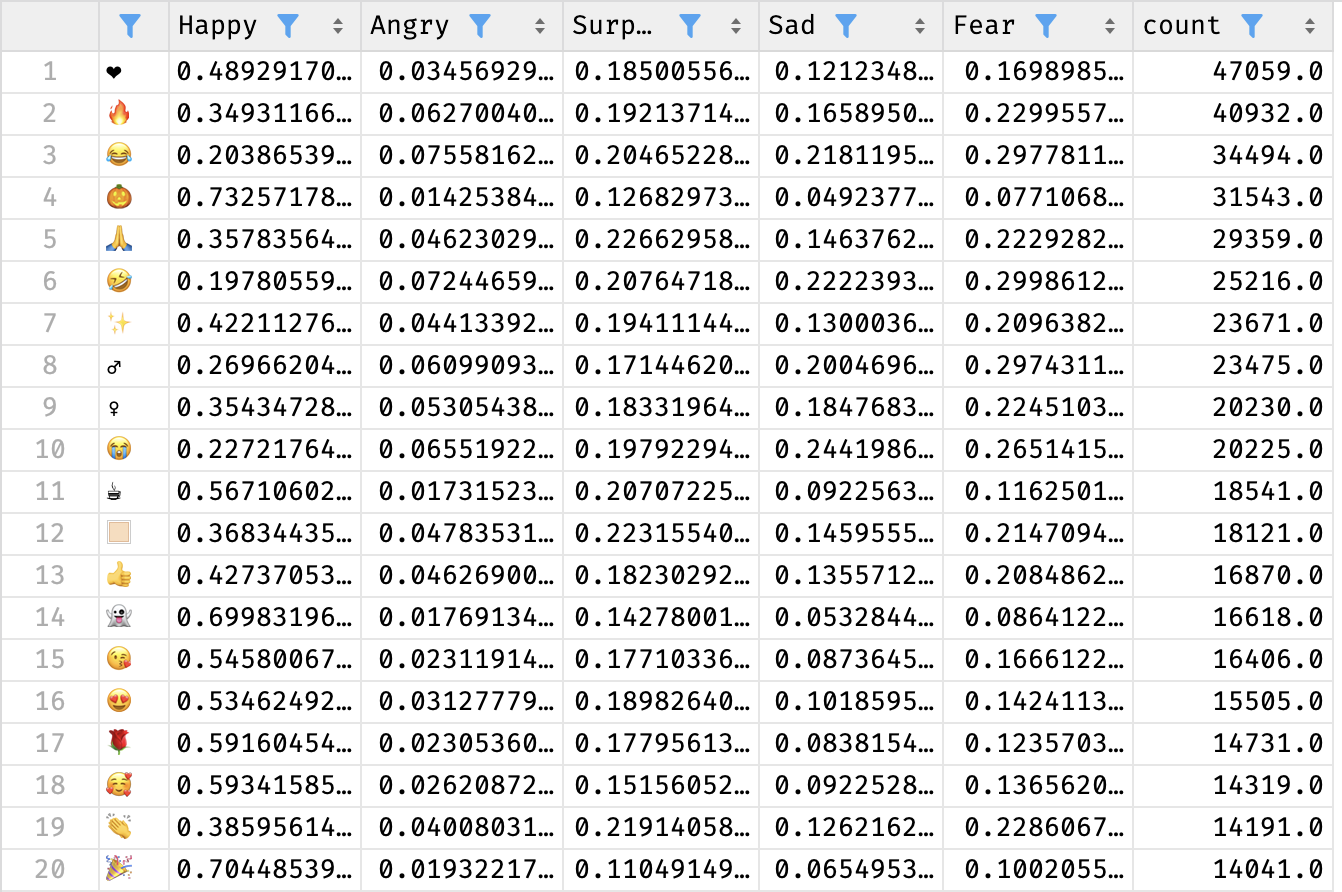}
    \caption{Emoji Emotion Ranking Dataset}
    \label{fig:emoji_emotion_dataset}
\end{figure}

\begin{figure}
    \centering
    \includegraphics[width=1\linewidth]{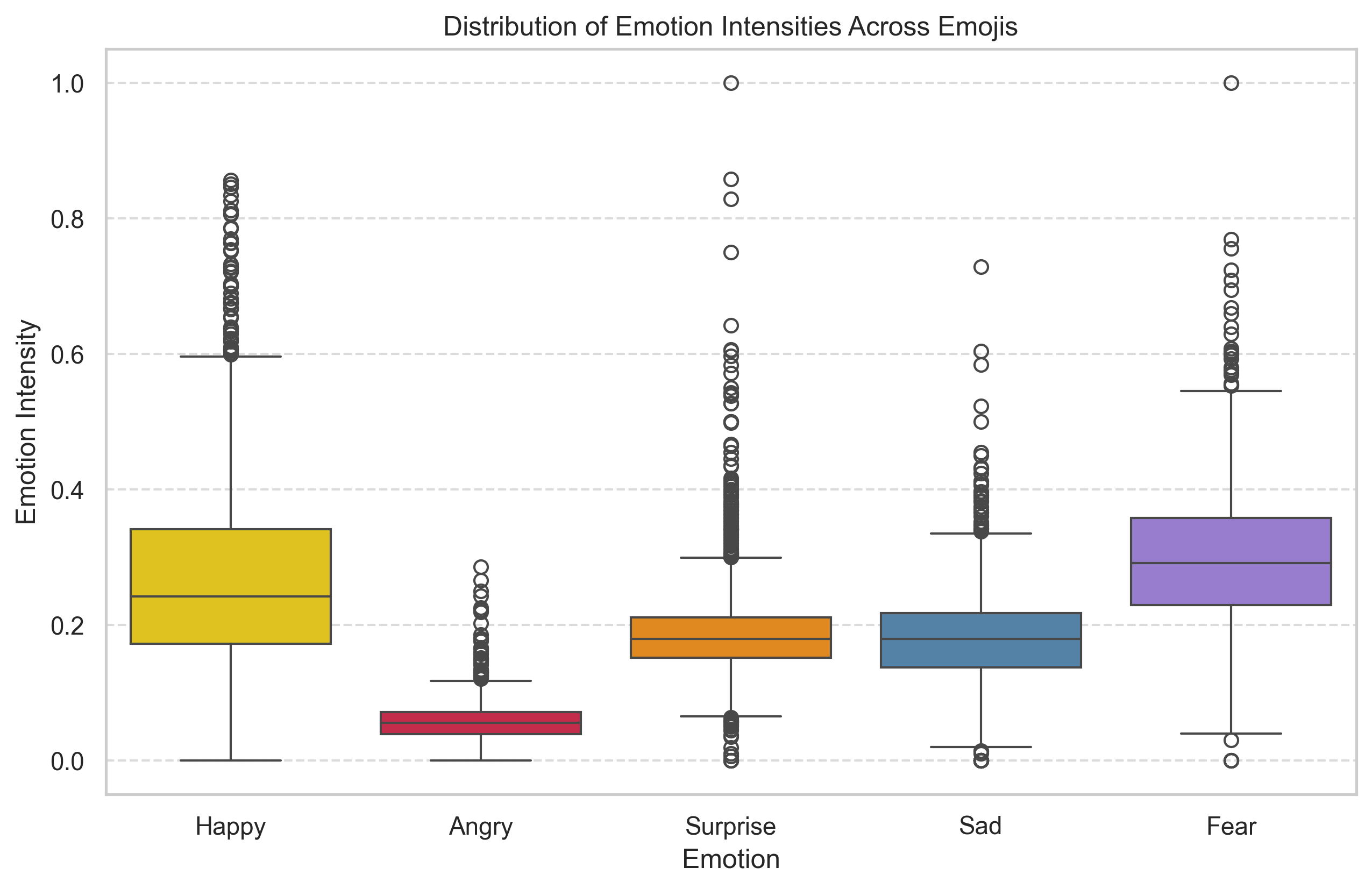}
    \caption{Emoji Emotion Box Plot}
    \label{fig:emoji_emotion_box_plot}
\end{figure}

Fig. \ref{fig:emoji_emotion_dataset} presents a sample of the constructed Emoji Emotion Ranking dataset generated. Each row corresponds to a unique emoji and includes normalized intensities for the five detected emotions: \textit{Happy}, \textit{Angry}, \textit{Surprise}, \textit{Sad}, and \textit{Fear}, along with the frequency of occurrence in the collected corpus. Fig. \ref{fig:emoji_emotion_box_plot} presents the distribution of emotion intensities across emojis in the dataset, visualized using box plots. Each box plot shows the distribution of emotion intensity values (0.0 to 1.0) for all emojis in the dataset. Fear exhibits the highest median intensity (approximately 0.30) and the widest distribution. In contrast, Angry shows the lowest median intensity (approximately 0.07) and the narrowest distribution.

\begin{figure}
    \centering
    \includegraphics[width=1\linewidth]{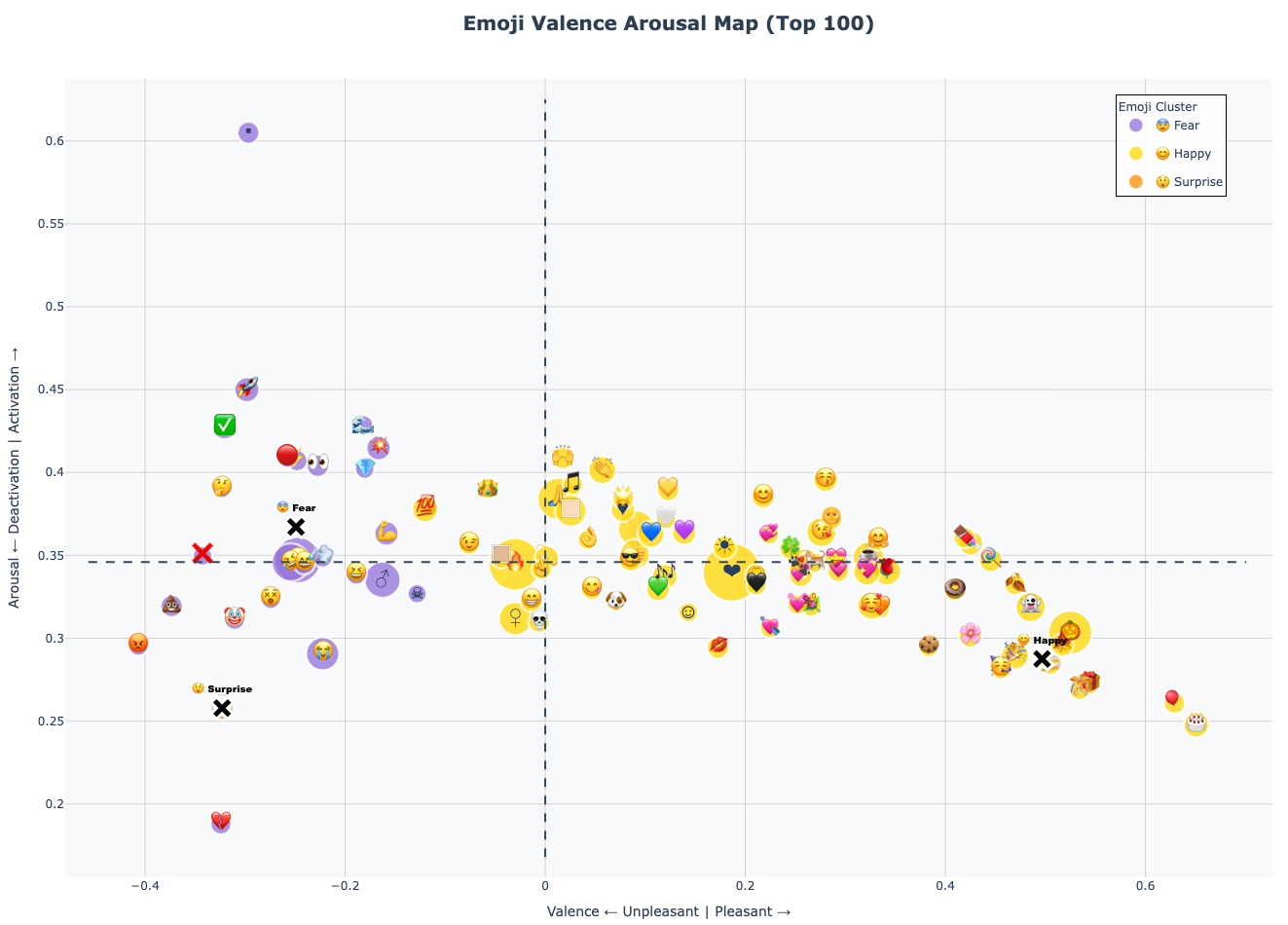}
    \caption{Valence Arousal Emoji Map}
    \label{fig:valence_arousal_emoji_map}
\end{figure}

Fig. \ref{fig:valence_arousal_emoji_map} illustrates the spatial distribution of emojis within a two-dimensional Valence–Arousal framework. In this representation, the horizontal axis corresponds to valence (ranging from negative to positive affect), while the vertical axis represents arousal (ranging from low to high emotional activation).

The coordinates assigned to each emoji were derived through an approximate mapping procedure based on the Russell Circumplex Model of Affect \cite{Russell1999}. 
The aggregated emotion scores associated with each emoji were then used to estimate its relative position in this two-dimensional space.

This projection enables the visualization of overall emotional tendencies of emojis in terms of both affective polarity (valence) and activation level (arousal). As a result, emojis expressing positive and highly activating emotions cluster in the upper-right quadrant, whereas negatively valenced and low-arousal emojis tend to appear in the lower-left region.


\begin{figure}[!t]
  \centering
  \includegraphics[width=1.0\linewidth]{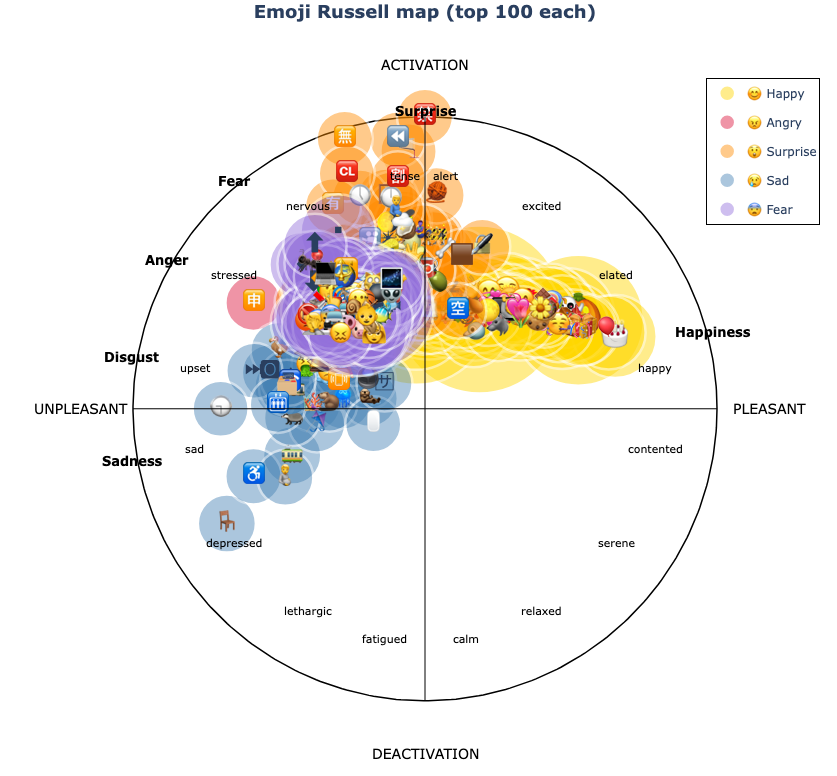}
  \caption{Emoji positions in the Russell circumplex (Valence on x, Arousal on y). Points are sized by frequency and colored by top emotion.}
  \label{fig:russell_emoji_map}
\end{figure}

Next, Fig. \ref{fig:russell_emoji_map} shows the distribution of emojis projected onto the Russell Circumplex Model of Affect, with valence on the horizontal axis and arousal on the vertical axis.

Each point in the figure corresponds to one emoji from the top 100 most frequent emojis in the dataset. The position of each emoji was determined by mapping its aggregated emotion intensities onto approximate valence and arousal coordinates derived from the theoretical placement of the five primary emotions (Happy, Fear, Sad, Surprise, and Angry) within the circumplex model, e.g., highly activating emotions (e.g., Angry or Surprise) are positioned higher on the arousal axis. The size of each point reflects the relative frequency of the emoji in the dataset, while colors indicate the dominant emotion associated with each emoji. 



\begin{figure} [tb]
    \centering
    \includegraphics[width=1\linewidth]{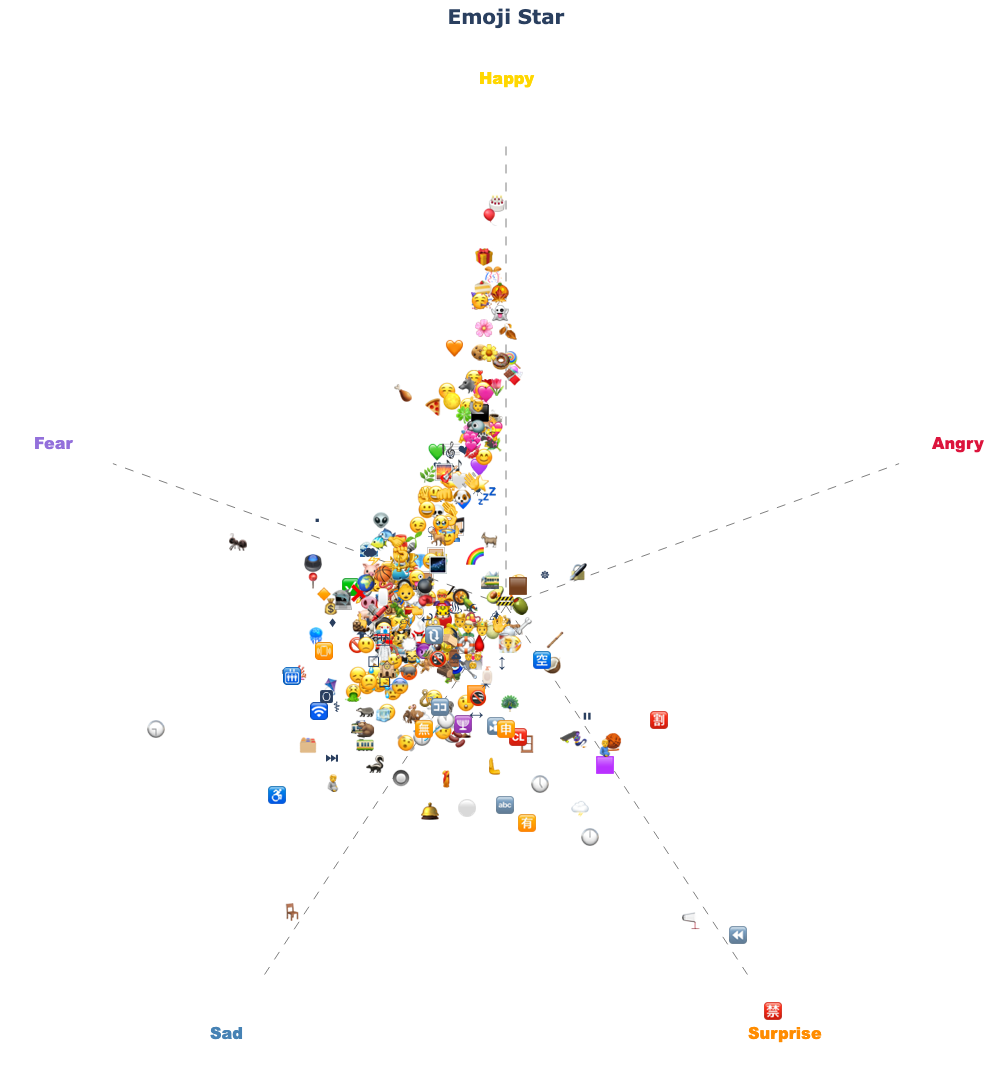}
    \caption{Emoji Emotion Star}
    \label{fig:emoji_emotion_star}
\end{figure}

Fig. \ref{fig:emoji_emotion_star} presents the emoji–emotion dataset visualized using a star (radar) diagram. In this representation, each axis of the star corresponds to one of the five primary emotions identified in the dataset: \textit{Happy}, \textit{Fear}, \textit{Sad}, \textit{Surprise}, and \textit{Angry}. 

For each emoji, the normalized emotion intensities are plotted along the respective axes, forming a polygon that characterizes its emotional profile. The distance from the center of the diagram indicates the relative intensity of a given emotion associated with the emoji. Larger radial values represent stronger emotional associations, whereas smaller values indicate weaker emotional contributions.



\begin{figure}
    \centering
    \includegraphics[width=1\linewidth]{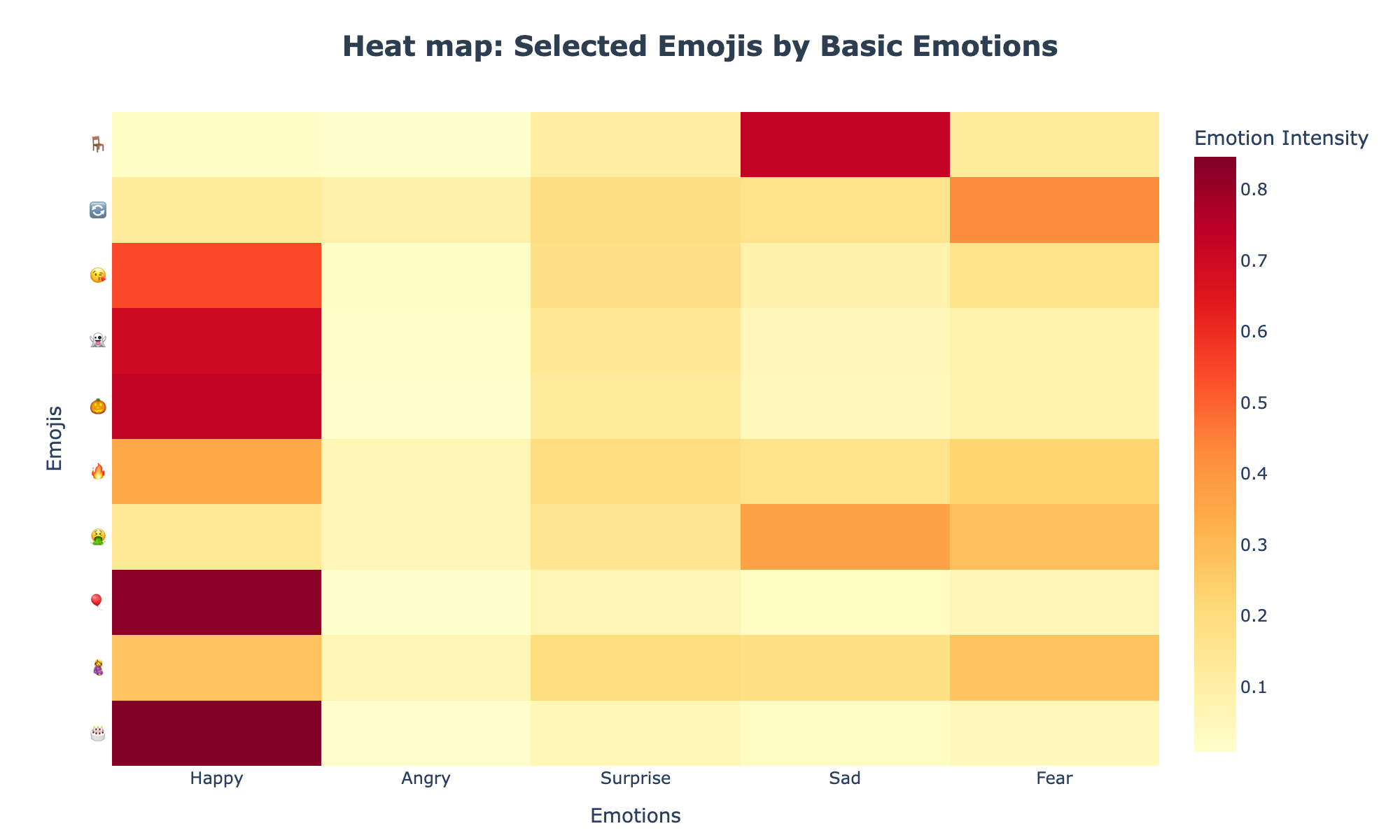}
    \caption{Emoji Emotion Heat Map}
    \label{fig:emojis_heatmap}
\end{figure}

Fig. \ref{fig:emojis_heatmap} presents a heatmap visualization of selected emojis and their associated emotion intensities. Each row in the heatmap corresponds to a specific emoji, while each column represents one of the five basic emotions. The color intensity of each cell indicates the strength of the emotional association between the corresponding emoji and emotion. This visualization enables intuitive comparison of emotional distributions across emojis. 







\section{Conclusion}
An emoji-aware framework was introduced that integrates dataset construction, emoji–emotion ranking, and projection into the Russell valence–arousal space. The study was conducted on a corpus of 100,000 emoji-containing Twitter (X) replies collected between 2020 and 2025, comprising 1,393 distinct emojis. By applying lexicon-based emotion detection and aggregating weighted emotion scores across contexts,  emoji-level emotional profiles were constructed and used to derive a systematic emoji–emotion ranking.

The results demonstrate that emojis exhibit multidimensional and probabilistic emotional characteristics rather than fixed sentiment polarity. The constructed ranking dataset revealed clear dominant emotional tendencies for frequently used emojis.


As for limitations, our dataset is restricted to English-language Twitter replies. It limits cross-cultural generalization. Future work may extend the framework to multilingual datasets and multiple social media networks. It can also explore the temporal dynamics of emoji usage. We also plan an extensive validation and comparison with other emotion-labeling approaches.






\bibliographystyle{IEEEtran}
\bibliography{references}


\balance
\end{document}